# Neuro-symbolic Models for Interpretable Time Series Classification using Temporal Logic Description

Ruixuan Yan*, Tengfei Ma†, Achille Fokoue†, Maria Chang†, Agung Julius*

* Department of Electrical, Computer, and Systems Engineering, Rensselaer Polytechnic Institute, Troy, NY, USA
{yanr5, juliua2}@rpi.edu

† IBM T.J. Watson Research Center, IBM Research, Yorktown Heights, NY, USA
{tengfei.ma1, maria.chang}@ibm.com, achille@us.ibm.com

*Abstract*—Most existing Time series classification (TSC) models lack interpretability and are difficult to inspect. Interpretable machine learning models can aid in discovering patterns in data as well as give easy-to-understand insights to domain specialists. In this study, we present Neuro-Symbolic Time Series Classification (NSTSC), a neuro-symbolic model that leverages signal temporal logic (STL) and neural network (NN) to accomplish TSC tasks using multi-view data representation and expresses the model as a human-readable, interpretable formula. In NSTSC, each neuron is linked to a symbolic expression, i.e., an STL (sub)formula. The output of NSTSC is thus interpretable as an STL formula akin to natural language, describing temporal and logical relations hidden in the data. We propose an NSTSC-based classifier that adopts a decision-tree approach to learn formula structures and accomplish a multiclass TSC task. The proposed smooth activation functions for wSTL allow the model to be learned in an end-to-end fashion. We test NSTSC on a real-world wound healing dataset from mice and benchmark datasets from the UCR time-series repository, demonstrating that NSTSC achieves comparable performance with the state-of-the-art models. Furthermore, NSTSC can generate interpretable formulas that match with domain knowledge.

## I. Introduction

Time series classification (TSC) is one challenging task in machine learning (ML) and data mining [1], [2]. Time-series data (TSD) exists in many aspects of our life, such as finance [3], clinical medicine [4], etc. Numerous TSC algorithms have been developed to extract hidden patterns from complicated and long sequences of TSD [5]. The interpretability of patterns extracted from TSD is particularly important to domain experts, especially for specialists in clinical medicine, geological science, and genetic science. Enriching the learning model to be intuitive for the users to understand the rules from the model and make better decisions is a critical task. For example, detection of seizures using data from Electroencephalograms (EEG) is a long-existing challenge, which has been tackled with various ML methods [6], [7], whereas there is no model that can simultaneously capture data patterns and describe the decisions as human-readable formulas.

Signal temporal logic (STL) is a formal language used to express or specify temporal specifications of cyber-physical systems [8], [9]. For instance, STL formulas can describe the features of the wound healing process as "*If the protein MMP.2's value is higher than $\underline{x}$ and lower than $\overline{x}$ during Day 1 to Day 10, then it's in the Hemostasis stage.*" Due to its expressivity, many *temporal logic inference (TLI)* algorithms have been proposed to analyze system executions by considering an STL formula as a classifier [10], [11]. Though the above models can learn STL formulas as a description of temporal properties of data, they face a computational complexity barrier, especially for long sequences of TSD. Classical *TLI* algorithms learn STL formulas by casting it as an optimization problem, incorporating the *quantitative satisfaction function* (will explain in Section V) into the objective function. Due to non-smoothness of the *quantitative satisfaction function*, the parameters are learned via expensive non-gradient-based methods. Moreover, the formulas from the above methods cannot reflect the relative importance of data at various time points because they model all time points uniformly.

In this study, we propose a novel differentiable Neuro-Symbolic framework for Time Series Classification (NSTSC) by using multi-view data representations, including the raw, spectral and derivative data. This framework associates a component of an STL formula with a neuron in a neural network (NN), making the model connected to an STL formula and also differentiable. Simultaneously, weights are assigned to the edges to reflect the impact of related inputs on a neuron's output, and the new form of STL is called weighted STL (wSTL). Bounded *quantitative satisfaction functions* for wSTL are proposed to reflect the truth degree of TSD with respect to a wSTL formula. The resulting wSTL formulas can therefore reflect the relative importance of data at various time points. Besides the raw data, we augment the data representation to obtain two additional data views: *spectral representation* and *derivative representation* (details to be explained in Seciton IV). As a result, NSTSC can provide wSTL formulas describing the characteristics of data in each view. Moreover, by utilizing a decision tree (DT) approach, the model's wSTL formula structures can be automatically learned and a multiclass TSC task can be accomplished. By traversing the paths of the tree classifier, individual wSTL formulas for each data class can be learned.

The contributions of this study are described as follows.

1) We propose novel weighted semantics for STL (called wSTL) using bounded *quantitative satisfaction functions* to quantify the truth degree of a wSTL formula over TSD. Any wSTL formula can be mapped to an NN, where a neuron is associated with a symbolic representation of a wSTL (sub)formula. In this way we make the model differentiable and easily optimized.

2) We develop a neuro-symbolic framework called NSTSC to classify TSD, using features that are expressed in wSTL. A decision tree approach is adapted to learn formula structures automatically and the resulted tree classifier can classify multiclass TSD, allowing the tree to provide wSTL formulas describing the characteristic patterns of each data class.

3) NSTSC is evaluated on a mouse wound-healing dataset and a set of benchmark time-series datasets from the UCR time series archive for comparison with state-of-the-art models. NSTSC is demonstrated to achieve competitive performance as well as express discriminatory patterns as interpretable logical formulas that match with domain knowledge.

## II. RELATED WORK

### A. Time Series Classification Models

Numerous TSC approaches have been developed to categorize TSD from various perspectives. Shapelet Transform (ST) [12] extracts shapelets, i.e., TSD subsequences, as features and classifies data based on their resemblance to the class representatives. ROCKET, the Random Convolutional Kernel Transform [13], applies random conventional kernels to TSD to extract summary statistics and utilizes a linear classifier for feature selection. In addition to the feature-based methods, neural networks (NNs) for TSC have recently been developed, such as residual network (ResNet), and fully convolutional NN (FCN) [14]. Ensemble learning models have also emerged to classify TSD. The Hierarchical vote Collective of Transformation-based Ensembles (HCTE) is an ensemble model that classifies data using a weighted average of predictions from its base classifiers: ST, Bag-Of-Symbolic Fourier-approximation-Symbols (BOSS), Time Series Forest (TSF), and Random Interval Spectral Ensemble (RISE). Recently, HCTE was upgraded to HCTE 2.0 [15] by replacing the base estimators with Temporal Dictionary Ensemble [16], Diverse Representation Canonical Interval Forest [17], and Arsenal [15]. The Time Series Combination of Heterogeneous and Integrated Embedding Forest (CHIEF) is an ensemble TSC model comprising trees as base estimators, which consider various features and splitting criteria for classification [18]. This work is concerned with designing a neuro-symbolic framework that can not only classify TSD but also express the model as an interpretable formula for better decision making.

### B. Signal Temporal Logic Inference

There has been a large body of STL inference works on TSD analysis during the last decades. The majority of them infer a formula by incorporating the *quantitative satisfaction function* into the objective function and solving an optimization problem. A parametric STL (pSTL) was proposed in [19] to find parameters for STL formulas. A CensusSTL describing the number of agents satisfying an STL formula in different groups was proposed in [20], accompanied with a CensusSTL inference algorithm. These methods have high computation costs due to the non-smooth objective functions and large search space. In this study, we introduce importance weights into STL to obtain differentiable *quantitative satisfaction functions* and develop a neuro-symbolic model that can learn the parameters of STL formulas in an end-to-end fashion.

### C. Neuro-Symbolic Models

In recent years, neuro-symbolic models have been extensively investigated in various disciplines [21]–[23]. Many-valued logics, such as fuzzy logic, have strong connections with wSTL and have been applied to time series forecasting [24]. However, the model has no relation to neural networks. Neuro-symbolic models seek to combine the benefits of robust learning in neural networks with the benefits of interpretability and reasoning in symbolic representations. Markov logic networks share a similar aim, using Markov networks to model logical formulas and to handle inference under uncertainty. Logical Neural Networks (LNN) [25] takes this a step further by using neural networks to model formulas in real-valued logics. LNNs have been applied to bi-directional inference and theorem proving tasks, as well as question answering over knowledge bases [26]. Recently, a rule-based representation learner (RRL) was developed in [22] to automatically learn data representation and classification as interpretable rules, where a gradient crafting approach is adapted to allow for continuous learning. Nevertheless, these models do not apply to TSD that has no grounding truth value.

## III. PRELIMINARIES

In this study, TSD refers to $p$-dimensional data that evolves in discrete time and is denoted as $x = \{x(0), x(1), ..., x(K-1)\}$, where $x(k) \in \mathbb{R}^p, \forall k < K$, $k$ is the time index, $K \in \mathbb{Z}_+$ is the duration of $x$, and $p \in \mathbb{Z}_+$. The TSC task is performed on a labeled dataset $D = \{(x_i, y_i)\}_{i=1}^N$, where $x_i$ is the $i$-th data in the dataset ($x_i = \{x_i(0), ..., x_i(K-1)\}, x_i(k) \in \mathbb{R}^p$), and $y_i$ is the label of $x_i$. The total number of classes in the dataset is denoted as $C$, meaning $y_i \in \mathcal{C} = \{1, 2, ..., C\}$.

### A. Weighted Signal Temporal Logic (wSTL)

The concept of wSTL was first introduced in [27]. In this study, we propose novel bounded quantitative semantics for wSTL based on the notion of truth degree.

**Definition 1.** *The syntax of wSTL is defined as [27]*

$$\phi := \top | \pi | \neg \phi | \phi_1^{w_1} \wedge \phi_2^{w_2} | \phi_1^{w_1} \vee \phi_2^{w_2} | \Diamond_{[k_1, k_2]}^{\boldsymbol{w}} \phi | \Box_{[k_1, k_2]}^{\boldsymbol{w}} \phi, \quad (1)$$

*where $\top$ is Boolean TRUE, $\pi$ is an atomic predicate defined as $\pi := f(x) \geq 0$. E.g., if $f(x) = \boldsymbol{a}^T x - u, \|\boldsymbol{a}\| = 1, u \in \mathbb{R}$, then $\pi$ represents a half space and $\boldsymbol{a}$, $u$ are parameters to learn. The symbols $\neg, \wedge$, and $\vee$ denote logical negation, conjunction, and disjunction, respectively. The temporal operators $\Diamond$ and*

representations are collected.

*2) Spectral Representation:* Studies have shown classifying TSD from the frequency domain can reveal discriminatory features invisible in the raw data [28]. Spectral data representation is obtained via fast Fourier transform (FFT) as follows:

$$\hat{x}(\omega) = \sum_{k=0}^{K-1} x(k)(e^{-\mathfrak{i}\frac{2\pi}{K}\omega k}), \ \omega = 0, ..., K-1, \\ = \hat{a}(\omega) + \mathfrak{i}\hat{b}(\omega), \qquad (2)$$

where $\mathfrak{i}$ is the imaginary unit. The power spectrum is then used to express the spectral representation, i.e., $\hat{x} = \{\hat{x}(0), ..., \hat{x}(K-1)\}$, where $\hat{x}(\omega) = \sqrt{\hat{a}^2(\omega) + \hat{b}^2(\omega)}$.

*3) Derivative Representation:* For derivative representation, we adopt the same derivative transformation technique as [29], which has shown the efficacy of derivative representation in TSC. The derivative representation is obtained via computing the first-order difference, i.e., $\tilde{x}(k) = x(k+1) - x(k)$.

### B. Interval Feature Extraction

Discriminatory interval features have been demonstrated to be effective in classifying TSD [28]. We split the TSD into intervals with length $\mathcal{I}$ and extract features from each interval. Specifically, given an aggregation function $f_A(\cdot)$ and an interval $[k \cdot \mathcal{I}, (k+1) \cdot \mathcal{I} - 1]$, $f_A(\cdot)$ is applied to the interval data $\{x(k \cdot \mathcal{I}), ..., x((k+1) \cdot \mathcal{I} - 1)\}$ to obtain an interval feature $x_{IF}(k) = f_A(x, k \cdot \mathcal{I}, (k+1) \cdot \mathcal{I} - 1)$. For example, if $f_A(\cdot) = \text{mean}(\cdot)$, $k = 0$, then the mean of the data within the interval $[0, \mathcal{I} - 1]$ is denoted as $x_{IF}(0) = \text{mean}(\{x(0), ..., x(\mathcal{I} - 1)\})$. By applying $f_A(\cdot)$ to the entire TSD, we can extract a sequence of interval features with length $M$, i.e., $x_{IF} = \{x_{IF}(0), ..., x_{IF}(M-1)\}$, where $M = \lceil \frac{K}{\mathcal{I}} \rceil$. For each data representation, the same interval feature extraction technique is applied. The interval feature of $\hat{x}$ (resp. $\tilde{x}$) is denoted as $\hat{x}_{IF}$ (resp. $\tilde{x}_{IF}$). The data representations and their corresponding interval features are given as inputs to the classifier, which learns a $\phi$ describing the characteristics of data and interval features in each view. Please see Fig. 3 for an illustration.

## V. NEURO-SYMBOLIC TIME SERIES CLASSIFICATION MODULE

The neuro-symbolic time series classification module relies on constructing a tree phase classifier (TPC) from the node phase classifiers (NPCs) for multiclass classification and formula structure learning. For clarity of the presentation, the variable notations of the NSTSC module is shown in Table I.

### A. Node Phase Classifier (NPC)

An NPC aims to design a neural network for a wSTL formula $\phi$, where each neuron is linked to a symbolic expression in a wSTL formula. This section first presents the activation functions (AFs), i.e., quantitative satisfaction functions, for the components in (1) such that every neuron can be mapped to a symbolic interpretation, then presents the design of an NPC using the AFs, and finally discusses the learning of an NPC.

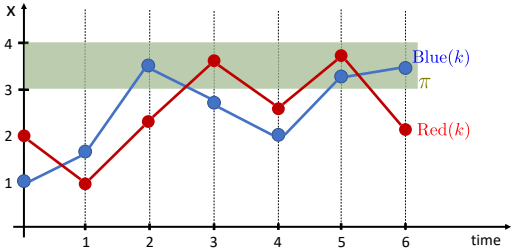

Fig. 1. Illustration of unweighted version of the STL semantics.

$\Box$ mean "Eventually" and "Always", respectively. We assume that $k_1 \leq k_2 < K$. The (unweighted) expression $\Diamond_{[k_1,k_2]}\phi$ means the formula $\phi$ is satisfied somewhere within the time interval $[k_1, k_2]$, while $\Box_{[k_1,k_2]}\phi$ means the formula $\phi$ is satisfied everywhere within the time interval $[k_1, k_2]$. See the example below for further explanation. $w_1$ and $w_2$ are nonnegative weights associated with $\phi_1$ and $\phi_2$ in the $\wedge$ and $\vee$ operators, and $\boldsymbol{w} = [w_{k_1}, w_{k_1+1}, ..., w_{k_2}]^T \in \mathbb{R}_{\geq 0}^{k_2-k_1+1}$ denotes nonnegative weights associated with $\Box$ and $\Diamond$ operators, with $w_{k'}$ being the weight assigned to time $k' \in [k_1, k_2]$.

**Example 1.** *To help understanding of the temporal logic semantics, we illustrate an example of the unweighted version of STL in Fig. 1. Two time series data, Red(k) and Blue(k), are shown. The green band represents a logical predicate $\pi$ : $(x \leq 4) \wedge (x \geq 3)$. An unweighted STL formula $\phi_1 \triangleq \Diamond_{[3,4]} \pi$ means the logical predicate $\pi$ is satisfied somewhere in the time interval [3,4]. Note that Red(k) satisfies $\phi_1$ while Blue(k) does not. We can thus use $\phi_1$ as a classifier between the two TSD. Similarly, an unweighted STL formula $\phi_2 \triangleq \Box_{[5,6]} \pi$ means the logical predicate $\pi$ is satisfied everywhere in the time interval [5,6]. We see that Blue(k) satisfies $\phi_2$ while Red(k) does not, implying that $\phi_2$ would also be a good classifier. The weights of the formulas, as we will explain later, are used to describe the relative importance of the time indices in the temporal operations or the clauses in the logical operations.*

## IV. MULTI-VIEW DATA REPRESENTATION & FEATURE EXTRACTION

**Model Structure Overview:** The overall structure of the TSC model is shown in Fig. 2. The raw TSD is first transformed into multi-view data representations, which are then used to extract the interval features. The multi-view data representations and the extracted interval features are given as inputs to the NSTSC. NSTSC relies on constructing a decision tree to classify the input data, where Node $i$ in the tree is a binary classifier designed as a neuro-symbolic model characterizing a wSTL formula $\phi_{Bi}$, which describes the characteristics of data and features in each view. Traversing the path from the root node to a leaf node gives the prediction of data and a formula for the data class in that leaf node.

### A. Multi-View Data Representation

*1) Raw Representation:* The original TSD is considered as the raw representation, from which spectral and derivative

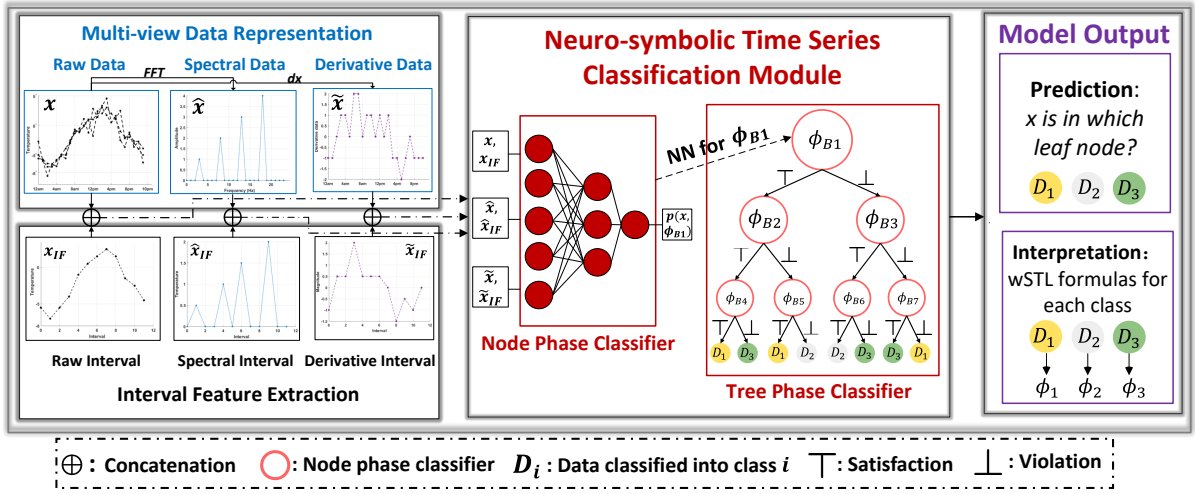

Fig. 2. The overall model structure of NSTSC, where the multi-view data and the extracted interval features are given as inputs to NSTSC, and the classification task is accomplished by learning a tree phase classifier. The leaf nodes represent data that are classified into a particular class. The tree phase classifier gives a wSTL formula describing each data class, e.g., the formula for data with label 2 is $\phi_2 = (\phi_{B1} \wedge \neg\phi_{B2} \wedge \neg\phi_{B5}) \vee (\neg\phi_{B1} \wedge \phi_{B3} \wedge \phi_{B6})$.

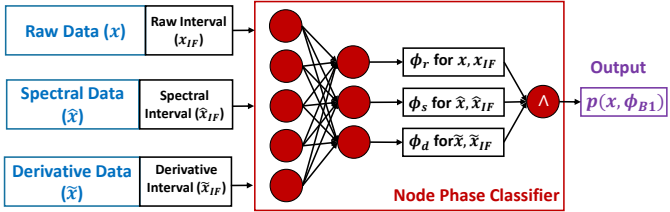

Fig. 3. Illustration of overall workflow of the node phase classifier for $\phi_{B1}$, which is in the form of $\phi_{B1} = \phi_r^{w_r} \wedge \phi_s^{w_s} \wedge \phi_d^{w_d}$. $\phi_r$, $\phi_s$, and $\phi_d$ describes the properties of data and features in the raw, spectral and derivative representations, respectively, and $w_r$, $w_s$, $w_d$ are the corresponding weights. The node phase classifier's output is the truth degree of $x$ with respect to $\phi_{B1}$, denoted as $p(x, \phi_{B1})$.

TABLE I
VARIABLE NOTATIONS OF THE NSTSC MODULE.

| Variable notation | Representation |
|---|---|
| $x$ / $\hat{x}$ / $\tilde{x}$ | raw / spectral / derivative data |
| $x_{IF}$ / $\hat{x}_{IF}$ / $\tilde{x}_{IF}$ | raw / spectral / derivative interval features |
| $\pi$ | atomic predicate |
| $\neg$ / $\wedge$ / $\vee$ | negation / conjunction / disjunction |
| $\Diamond$ / $\Box$ | "Eventually" / "Always" |
| $p(x, \phi, k)$ | truth degree of $\phi$ over $x$ at $k$ |
| $\phi_r, \phi_s, \phi_d$ | subformula describing the raw, spectral, derivative data |
| $\mathcal{B}$ | formula structure base |
| $D^n$ | data assigned to Node $n$ |
| $D_c^n$ | data in Node $n$ with encoded label for Class $c$ |
| $\phi_B^c$ | formula from an NPC for class $c$ with structure $\phi_B \in \mathcal{B}$ |

**Definition 2.** *To describe the truth degree, i.e., the degree of satisfaction of $\phi$ over $x$ at $k$, denoted as $p(x, \phi, k)$, we define the bounded quantitative satisfaction functions for wSTL as*

$$p(x, \pi, k) = g(f(x(k))), \quad (3)$$
$$p(x, \neg\phi, k) = 1 - p(x, \phi, k),$$
$$p(x, \phi_1^{w_1} \wedge \phi_2^{w_2}, k) = \otimes^{\wedge}([w_j, p(x, \phi_j, k)]_{j=1,2}),$$
$$p(x, \phi_1^{w_1} \vee \phi_2^{w_2}, k) = \oplus^{\vee}([w_j, p(x, \phi_j, k)]_{j=1,2}),$$

$$p(x, \Diamond^{\bm{w}}\phi, k) = \oplus^{\Diamond}([w_{k'}, p(x, \phi, k+k')]_{k' \in [k_1, k_2]}),$$
$$p(x, \Box^{\bm{w}}\phi, k) = \otimes^{\Box}([w_{k'}, p(x, \phi, k+k')]_{k' \in [k_1, k_2]}),$$

*where $\pi := f(x) \geq 0$, $g(\cdot)$, $\otimes^{\wedge}(\cdot)$, $\oplus^{\vee}(\cdot)$, $\otimes^{\Box}(\cdot)$, $\oplus^{\Diamond}(\cdot)$ are activation functions for predicates and $\wedge, \vee, \Box, \Diamond$ operators, respectively. Note that (3) returns a value between 0 and 1 that represents the truth degree, different from the traditional quantitative satisfaction functions in, e.g., [10] that return values in $\mathbb{R}$. We denote $p(x, \phi, 0)$ as $p(x, \phi)$ for simplicity.*

*1) Design of Activation Functions for wSTL:* The bounded quantitative satisfaction functions should be such that $p(x, \phi, k)$ close to 1 (resp. 0) indicates that $x$ robustly satisfies (resp. violates) $\phi$ at $k$. In this study, we define $g(\cdot)$ in (3) as

$$g(r) = sigmoid(r) \triangleq \frac{1}{1 + e^{-r}}. \quad (4)$$

Note that $g(r)$ is differentiable everywhere with nonnegative derivative for any $r \in \mathbb{R}$. Essentially, temporal operators $\Box$ (resp. $\Diamond$) can be expressed as a sequence of logical operators $\wedge$ (resp. $\vee$). For instance, $\Box_{[k_1, k_2]}^{\bm{w}} \phi$ can be expressed as

$$\Box_{[k_1, k_2]}^{\bm{w}} \phi = \phi_{k_1}^{w_{k_1}} \wedge \phi_{k_1+1}^{w_{k_1+1}} \wedge \cdots \wedge \phi_{k_2}^{w_{k_2}}, \quad (5)$$

where $\phi_{k'}$ denotes the formula $\phi$ evaluated on data instance $x(k')$ for $k' \in [k_1, k_2]$. As a result, our task is to design activation functions (AFs) for the logical operators $\wedge$ and $\vee$. The AF for the $\vee$ operator can be derived from the AF for the $\wedge$ operator using the De Morgan's law, $\neg(\phi_1 \vee \phi_2) = (\neg\phi_1) \wedge (\neg\phi_2)$.

The AF for the $\wedge$ operator is defined as [25]:

$$\otimes^{\wedge}([w_j, p(x, \phi_j, k)]_{j=1,2}) \triangleq h(\beta - \sum_{j=1}^{2} \bar{w}_j(1 - p(x, \phi_j, k))), \quad (6)$$

where $\bar{w}_j = w_j/(w_1 + w_2)$ is the normalized weight, $w_j$ and $\beta$ are parameters to be learned, and $h(z) \triangleq \max\{0, \min\{z, 1\}\}$ is introduced to clamp the truth degree to the range $[0, 1]$.

We note that the defined AF satisfies the following properties (presented without proof due to space limitation).
- **Nonimpact for zero weights**: Truth degree of subformulas with zero weights do not impact the overall truth degree. Mathematically, if $w_j = 0$, then $p(x, \phi_j, k)$ has no impact on $\otimes^\wedge([w_j, p(x, \phi_j, k)]_{j=1,2})$.
- **Impact ordering**: The impact of truth degree of subformulas on the overall truth degree follows the order of their weights. Mathematically, it means if $p(x, \phi_1, k) = p(x, \phi_2, k)$ and $w_1 \geq w_2$, then

$$\frac{\partial \otimes^\wedge([w_j, p(x,\phi_j,k)]_{j=1,2})}{\partial p(x,\phi_1,k)} \geq \frac{\partial \otimes^\wedge([w_j, p(x,\phi_j,k)]_{j=1,2})}{\partial p(x,\phi_2,k)}.$$

- **Monotonicity**: The value of the AF increases monotonically with respect to the truth degree of each clause. Mathematically, this property can be expressed as

$$\otimes^\wedge([w_j, p(x, \phi_j, k)]_{j=1,2}) \leq \otimes^\wedge([w_j, p(x, \phi_j, k) + d]_{j=1,2}), d \geq 0.$$

By De Morgan's law, the activation function for $\vee$ is

$$\oplus^\vee([w_j, p(x, \phi_j, k)]_{j=1,2}) \triangleq h(1 - \beta + \sum_{j=1}^{2} \bar{w}_j p(x, \phi_j, k)). \quad (7)$$

By using the relationship in (5), the AFs for $\Diamond$ and $\Box$ are

$$\oplus^\Diamond([w_{k'}, p(x, \phi, k + k')]_{k' \in [k_1, k_2]}) \triangleq \quad (8)$$
$$h(1 - \beta + \sum_{k' \in [k_1, k_2]} \bar{w}_{k'} p(x, \phi, k + k') \mathbb{1}(k + k' < K)),$$

$$\otimes^\Box([w_{k'}, p(x, \phi, k + k')]_{k' \in [k_1, k_2]}) \triangleq \quad (9)$$
$$h(\beta - \sum_{k' \in [k_1, k_2]} \bar{w}_{k'} (1 - p(x, \phi, k + k')) \mathbb{1}(k + k' < K)),$$

where $\bar{w}_{k'} = w_{k'} / \sum_{k'' \in [k_1, k_2]} w_{k''}$, $h(z) \triangleq \max\{0, \min\{z, 1\}\}$ is to clamp the truth degree to the range $[0, 1]$, $\mathbb{1}$ is the indicator function for ignoring the impact of subformulas beyond $K - 1$ (the length of the data).

*2) Design of an NPC:* With the AFs in (4) - (9), a neuro-symbolic model can be built for any $\phi$, which is illustrated using Example 2.

**Example 2.** *An example of the NN design for $\phi = \Box_{[0,2]}^{w^2}(\Diamond_{[0,2]}^{w^1}\pi)$ is shown in Fig. 4. The first layer is the input, the second layer represents the predicate $\pi$, the third layer represents the $\Diamond_{[0,2]}$ symbol, and the fourth layer represents the $\Box_{[0,2]}$ symbol. The NN's output is the truth degree $p(x, \phi)$.*

Fig. 4 implies an NN's structure at a node depends on the structure of $\phi$, i.e. the arrangement of operators and predicates in $\phi$. In this study, the formula structure of an NPC is chosen from a formula structure base denoted as $\mathcal{B}$.

**Formula Structure Base:** Each node considers a set of commonly used formula structures in STL inference works as the structure base [30], which is described as follows:
1) Distinct conjunctive or disjunctive patterns:

$$\mathcal{B}_1 = \{\phi_0^{w_0} \wedge ... \wedge \phi_{K-1+M}^{w_{K-1+M}} \text{ or } \phi_0^{w_0} \vee ... \vee \phi_{K-1+M}^{w_{K-1+M}}\},$$

where $\phi_k$ is a predicate describing the $k$-th input data, $M$ is the number of interval features.

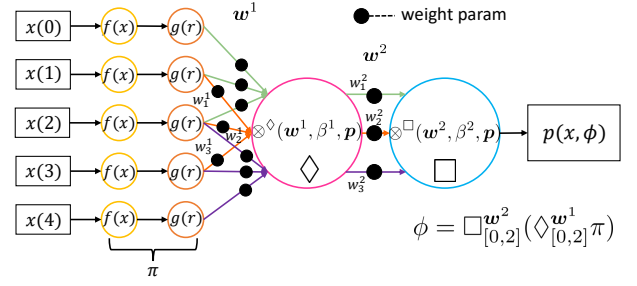

Fig. 4. NN structure of an NPC for the wSTL formula $\phi = \Box_{[0,2]}^{w^2}(\Diamond_{[0,2]}^{w^1}\pi)$.

2) Consistent or alternative pattern:

$$\mathcal{B}_2 = \{\Box_{[0, K-1+M]}^{w}\phi \text{ or } \Diamond_{[0, K-1+M]}^{w}\phi\}. \quad (10)$$

$\phi$ is a common predicate applying to any $k \in [0, K - 1 + M]$.
3) Persistent or eventually consistent pattern:

$$\mathcal{B}_3 = \{\Box_{[0, K-1+M]}^{w^1}\Diamond_{[0, K-1+M]}^{w^2}\phi \text{ or } \Diamond_{[0, K-1+M]}^{w^1}\Box_{[0, K-1+M]}^{w^2}\phi\},$$

where $\phi$ is a common predicate similar to the one in (10).
The formula structure base is denoted as $\mathcal{B} \triangleq \mathcal{B}_1 \cup \mathcal{B}_2 \cup \mathcal{B}_3$.

**Remark 1.** *The formula structure base defined above is not the only choice. Users can define other types of formula structure bases such as more general ones or specific ones guided by prior knowledge or domain knowledge on data.*

*3) Learning of an NPC:* Given the data $D^n$ of Node $n$ and a basic formula structure $\phi_B \in \mathcal{B}$, a node classifier for class $c \in \mathcal{C}$ builds an NSTSC for $\phi_B$ and classifies each data in $D^n$ as either an instance of class $c$ or not. Each node classifier is essentially a binary classifier that is trained to learn a formula for a particular class. While classifying the data with label $c$ and the remaining data, the data labels in $D^n$ are one-hot encoded via $\hat{y}_i^c = \mathbb{1}(y_i = c)$, where $\mathbb{1}$ is the indicator function. In doing so, we could obtain $\mathcal{D}_c^n = \{(x_i, \hat{y}_i^c)|(x_i, y_i) \in D^n\}$ representing the data assigned to Node $n$ with encoded label for class $c$. A node classifier outputs $p(x, \phi_B) \in [0, 1]$, and $p(x, \phi_B) \geq 0.5$ represents $x$ satisfies $\phi_B$, and $p(x, \phi_B) < 0.5$ represents $x$ violates $\phi_B$. As illustrated in Fig. 3, $\phi_B$ is in the form of $\phi_B = \phi_r^{w_r} \wedge \phi_s^{w_s} \wedge \phi_d^{w_d}$, where $\phi_r, \phi_s, \phi_d$ describes raw, spectral, and derivative data and interval features, respectively, and $w_r, w_s, w_d$ are the corresponding weights. $\phi_r, \phi_s, \phi_d$ take the formula structures in $\mathcal{B}$. The loss function is defined as the standard weighted cross-entropy loss:

$$L(\mathcal{D}_c^n, \phi_B) = \sum_{i=1}^{|\mathcal{D}_c^n|} (-I_R \hat{y}_i^c \log p(x_i, \phi_B) \quad (11)$$
$$- (1 - \hat{y}_i^c) \log(1 - p(x_i, \phi_B))),$$

where $I_R = |\{i|\hat{y}_i^c = 0\}| / |\{i|\hat{y}_i^c = 1\}|$, and $|\cdot|$ denotes the set cardinality. $I_R$ is thus the imbalance ratio introduced to tackle the data imbalance issue, which may happen when posing a multiclass classification problem as multiple binary classification problems. The learning of NSTSC is similar to the tradition NNs via back-propagation, which is described

**Algorithm 1** Node Formula Learning Algorithm - *NFLA()*

---
**Input:** $\mathcal{D}_c^n$: TSD assigned to Node $n$ with encoded label for class $c$, $\phi_B$: a basic formula in $\mathcal{B}$, $I_t$: maximum number of iterations
**Output:** Learned wSTL formula $\phi_B$ for an NPC
1: Construct an NSTSC based on the structure of $\phi_B$, and initialize the parameters $\boldsymbol{w}, \beta, \boldsymbol{a}, u$ in $\phi_B$
2: **for** $t = 1, 2, ..., I_t$ **do**
3:     Select a mini-batch data $\mathcal{D}_c^{n,t}$ from $\mathcal{D}_c^n$
4:     Run forward-propagation for NSTSC to compute the truth degree of $\phi_B$ over $\mathcal{D}_c^{n,t}$ using (4) - (9)
5:     Compute the loss at the current iteration using (11)
6:     Run back-propagation to update the parameters including $\boldsymbol{w}, \beta, \boldsymbol{a}, u$ in $\phi_B$
7: **end for**
8: **return** $\phi_B$

---

as Algorithm 1. The forward propagation computes the truth degree of $\phi_B$ over $x$, and the backward propagation updates the parameters in $\phi_B$. When back propagate to the neurons for the operators, $\boldsymbol{w}$ and $\beta$ in (6)-(9) are updated, and when back propagate to the neurons for the predicates $\pi$, the parameters defining $f(x)$ are updated. For instance, if $f(x) = \boldsymbol{a}^T x - u$, then the parameters $\boldsymbol{a}$ and $u$ are updated.

### B. Tree Phase Classifier (TPC)

While the expressiveness of $\phi_B \in \mathcal{B}$ is limited, combining multiple basic formulas could enrich the expressiveness. An NPC can complete a binary classification task. Multiclass TSC can be accomplished by constructing a decision tree (DT) [30] from the NPCs as a tree phase classifier (TPC). The procedures for constructing the TPC are presented in Algorithm 2, from which we could learn a formula $\phi_c$ for every data class $c$. Initially, we check if Node $n$ is the root node. If Node $n$ is the root node, then we set $\phi_{pt} = \emptyset$ (line 2). Next, we check if the stop condition is satisfied, e.g. $h \leq 5$ (line 4 - 6). If the stop condition is satisfied, then the tree up to the current node is returned as the tree classifier $\mathcal{T}$. Otherwise, we will train the node classifier for Node $n$. For each class $c \in \mathcal{C}$, we encode the labels of data in $D^n$ for class $c$ and obtain $\mathcal{D}_c^n$ (line 8). Then for each $\phi_B \in \mathcal{B}$, we use the given $\phi_B$ and the encoded dataset $\mathcal{D}_c^n$ to train an NPC for class $c$ via Algorithm 1. By doing so, we learn a formula $\phi_B^c$ that is with formula structure $\phi_B$ and can classify data with label $c$ and the remaining data in $D^n$. Next, we evaluate the performance of $\phi_B^c$ using the branching criterion defined as the standard Gini index, i.e.,

$$J(D^n, \phi_B^c) = \frac{|D_\top^n(\phi_B^c)|}{|D^n|} G_I(D_\top^n(\phi_B^c)) + \frac{|D_\bot^n(\phi_B^c)|}{|D^n|} G_I(D_\bot^n(\phi_B^c)), \quad (12)$$

where $D_\top^n(\phi_B^c)$ denotes the data in $D^n$ satisfying $\phi_B^c$, and $D_\bot^n(\phi_B^c)$ denotes the data in $D^n$ violating $\phi_B^c$, and

$$G_I(D_\top^n(\phi_B^c)) = 1 - \sum_{m=1}^{C} (p(D_\top^n(\phi_B^c), m))^2,$$

$$p(D_\top^n(\phi_B^c), m) = \frac{|\{i|(x_i, y_i) \in D_\top^n(\phi_B^c), y_i = m\}|}{|D_\top^n(\phi_B^c)|}$$

denotes the fraction of data in $D_\top^n(\phi_B^c)$ with original label $m$,

$$G_I(D_\bot^n(\phi_B^c)) = 1 - \sum_{m=1}^{C} (p(D_\bot^n(\phi_B^c), m))^2,$$

$$p(D_\bot^n(\phi_B^c), m) = \frac{|\{i|(x_i, y_i) \in D_\bot^n(\phi_B^c), y_i = m\}|}{|D_\bot^n(\phi_B^c)|}$$

denotes the fraction of data in $D_\bot^n(\phi_B^c)$ with original label $m$ (line 10). By repeating this step for each $\phi_B \in \mathcal{B}$ and each data class $c$, we could learn a set of formulas with different structures for each class $c$, from which we choose the formula $\phi_B^c$ with the best criterion as $\phi^{n,*}$ that represents the optimal formula at Node $n$ (line 13). Thus a greedy search strategy is utilized to learn the formula at a node. With $\phi^{n,*}$, $D^n$ is partitioned into $D_\top^n(\phi^{n,*})$ satisfying $\phi^{n,*}$ and $D_\bot^n(\phi^{n,*})$ violating $\phi^{n,*}$, which is denoted as $partition(D^n, \phi_{pt} \wedge \phi^{n,*})$ in line 14. Next, we generate two child nodes of $n$, denoted as Node $n+1$ and $n+2$, and distribute the data $D_\top^n(\phi^{n,*})$ and $D_\bot^n(\phi^{n,*})$ to Nodes $n+1$ and $n+2$, respectively (line 15). The *DTFL* algorithm is then performed on the two child nodes (line 16-17), where *n.left* and *n.right* denote the left and right child node of Node $n$, respectively. The above procedures proceed until the stop conditions are satisfied. The conversion from a TPC to a wSTL formula $\phi_c$ that describes the data with label $c$ can be completed by a similar approach to [30]. The details of the conversion are presented in Appendix A.

## VI. EXPERIMENTS

We evaluate NSTSC using two groups of datasets. One group is a set of real-world wound healing data from experiments on mice, and another is a set of datasets from the UCR TSD archive [31]. The experiments are run using the AdamW optimizer in Pytorch (1.10.2) on a macOS 11.4 system with a Quad-Core CPU (i7, 2.9GHz) and a 16GB RAM. Our code is available at https://tinyurl.com/NSTSC.

### A. Wound Healing Dataset

The wound healing data comes from real-world mice experiments. The task is to analyze and predict wound healing stages so that proper interventions can be done to accelerate the healing process.

*1) Dataset Description:* The wound healing process is divided into stages of hemostasis (Hem), inflammation (Inf), proliferation (Pro), and maturation (Mat). The dataset contains 67 mice data, each with two recording time points corresponding to the first day and the date of a healing stage. Every data has four features representing four protein levels: MMP.2 ($x^1$), IL.6 ($x^2$), PLGF.2 ($x^3$), and VEGF ($x^4$). We choose the following baseline models for comparison, including FCN, ResNet, and multi-layer perceptron (MLP) for TSC [14].

**Algorithm 2** Decision Tree Formula Learning - *DTFL()*

**Input:** $\phi_{pt}$: formula associated with the current path; $n$: the node index; $h$: the current depth; $\mathcal{B}$: formula structure base, $stop$: stop condition; $J$: branching criterion; $D^n$: TSD at Node $n$

**Output:** The tree phase classifier $\mathcal{T}$ for multiclass TSC

1: **if** $n = 0$ **then**
2:     Set $\phi_{pt} = \emptyset$
3: **end if**
4: **if** $stop(\phi_{pt}, n, h, D^n)$ **then**
5:     Return $\mathcal{T}$ as the multiclass TPC
6: **end if**
7: **for** $c \in \mathcal{C} = \{1, 2, ..., C\}$ **do**
8:     Encode labels of data in $D^n$ for class $c$ and obtain $\mathcal{D}_c^n$
9:     **for** $\phi_B \in \mathcal{B}$ **do**
10:         Run Algorithm 1 using $\phi_B$ and $\mathcal{D}_c^n$ to learn $\phi_B^c$, compute $J(\mathcal{D}^n, \phi_B^c)$ using (12)
11:     **end for**
12: **end for**
13: $\phi^{n,*} = \arg\min_{\phi_B^c} J(\mathcal{D}^n, \phi_B^c)$
14: $n \leftarrow \phi^{n,*}, D_\top^n(\phi^{n,*}), D_\bot^n(\phi^{n,*}) \leftarrow partition(D^n, \phi_{pt} \wedge \phi^{n,*})$
15: $D^{n+1} \leftarrow D_\top^n(\phi^{n,*}), D^{n+2} \leftarrow D_\bot^n(\phi^{n,*})$
16: $n.left \leftarrow DTFL(\phi_{pt} \wedge \phi^{n,*}, n + 1, h + 1, \mathcal{B}, stop, J, D^{n+1})$
17: $n.right \leftarrow DTFL(\phi_{pt} \wedge \neg\phi^{n,*}, n + 2, h + 1, \mathcal{B}, stop, J, D^{n+2})$

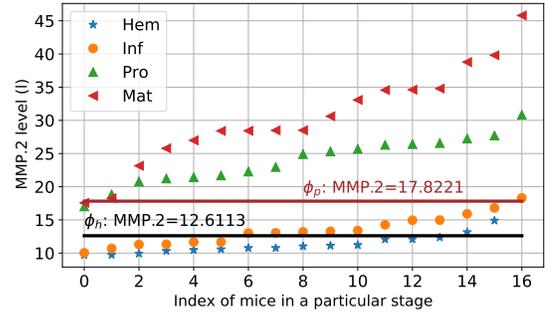

Fig. 5. Expression of "MMP.2" for four stages and decision rules in $\phi_h$ (black line) and $\phi_p$ (brown line).

*2) Experiment Setup and Results:* As data is given as four proteins, we consider using a predicate in the form of $\pi_j := a^j x^j - u^j \geq 0$ for each protein, where $x^j$ denotes the $j$-th protein feature, and $a^j$ and $u^j$ are parameters to learn. The interval features are omitted for the wound healing dataset as the sequence is short. The formula structure is chosen from $\phi = \phi_1^{w_1} \wedge \phi_2^{w_2}$ and $\phi = \phi_1^{w_1} \vee \phi_2^{w_2}$, where $\phi_1$ describes data on the first day, and $\phi_2$ describes data on the date of a specific stage. The task for the wound healing dataset is to predict the healing stage for a mouse given its wound data. A stratified 5-fold cross validation approach is used for data split. The parameters $a^j, u^j$ are initialized as $1 \times 10^{-5}$, and $w$ are initialized as random numbers from a uniform distribution in the range of $[0, 1)$, $\beta$ are initialized as 1. The maximum epoch number is 100, and the learning rate (LR) is 0.1. The mean and standard deviation (std) of accuracy, recall, precision, and F1 score are shown in Table II. We can observe that NSTSC achieves better results than the baseline models regarding higher classification accuracy and F1 score.

*3) Interpretability Analysis:* Studies have shown that MMP.2 level will increase from the Hem stage to the Pro stage [32]. The formula learned by NSTSC for the Hem stage is $\phi_h = (\phi_h^1)^{0.0143} \wedge (((\boldsymbol{x^1 \leq 12.6113})^{0.9593} \wedge (x^2 \leq 0)^{0.0355} \wedge (x^3 \geq 0.9196)^{0.0051})^{0.99} \vee (x^4 \geq 0)^{0.01})^{0.9857}$, and the formula learned for the Pro stage is $\phi_p = (\phi_p^1)^{0.0847} \wedge (((\boldsymbol{x^1 \geq 17.8221})^{0.5740} \wedge (x^2 \geq 0)^{0.3950} \wedge (x^3 \leq 0.0367)^{0.0309})^{0.9737} \vee (x^4 \geq 0)^{0.0263})^{0.9153}$, where $\phi_h^1$ and $\phi_p^1$ describe data on the first day, and the value of the theresholds in the predicates are all learned by the model. If we extract the important subformulas, $\phi_h$ reads as "MMP.2 is below 12.6113 at $k = 2$", and $\phi_p$ reads as "MMP.2 is above 17.8221 at $k = 2$". This identifies "MMP.2" increases from the Hem stage to the Pro stage, matching the domain knowledge of MMP.2's increase in [32]. Also, we visualize "MMP.2" level at $k = 2$ and the decision rules of $\phi_h$ ($l \leq 12.6113$) and $\phi_p$ ($l \geq 17.8221$) in Fig. 5, where x-axis denotes the index of mice belonging to a particular stage, and y-axis denotes the "MMP.2" level ($l$). We can observe "MMP.2" of most mice in the Hem stage is below 12.6113, and "MMP.2" of most mice in the Pro stage is above 17.8221, which matches with the patterns described by $\phi_h$ and $\phi_p$. This demonstrates NSTSC can provide interpretable rules that match with domain knowledge and can assist clinical experts in making better decisions on the wound healing stage prediction. Although the wound healing data has only two time points, NSTSC is proven to provide clinical experts with practically useful rules that are human-readable and easy to understand. In addition, we use the following UCR dataset to show that NSTSC also works for long sequences of TSD.

### B. UCR Time-Series Data

We implement NSTSC on the UCR TSD archive [31] for the second experiment. The standard split of training data and test data is utilized here. The $f(x)$ in the atomic predicates are set in the form of $f(x) = \boldsymbol{a}^T x - u, \|\boldsymbol{a}\| = 1$. The parameters defining the atomic predicates are initialized as $1 \times 10^{-5}$. The weights for operators $\boldsymbol{w}$ are initialized as random numbers from a uniform distribution in the range of $[0, 1)$, and $\beta$ are initialized as 1, and the LR is set as 0.1. The number of training epochs is set as 100. We compare NSTSC with the following state-of-the-art (SOTA) models: FCN and ResNet [14], ST [12], BOSS [33], HCTE [28], HCTE2 [15], ROCKET [13], and CHIEF [18]. 85 commonly used datasets in the above SOTA works are selected for comparison purposes.

*1) Experimental Results:* In the experiment, the stop condition depends on the maximal depth of the DT. The maximal tree depth is five for the binary datasets, seven for the multiclass datasets with less than 10 classes, and nine for the other datasets. Similar to [34], NSTSC adopts the mean, variance, maximum, minimum, interquartile range, and slope aggregation function to extract interval features. The number of intervals is set as 20, and the interval length $\mathcal{I}$ is determined correspondingly. We report the accuracy on the test data using the model with the best criterion on the test data. The number of first place rank among all systems (No. rank 1st) and the average accuracy for all the datasets and for only binary datasets are shown in Table III and Table IV, respectively. As shown in Table III, NSTSC achieves an average accuracy of $84.66\%$, which is only lower than HCTE, HCTE2, CHIEF, and ROCKET. However, the gap between NSTSC and the other models is at most $1.46\%$. Moreover, NSTSC is ranked first more often than the other systems except HCTE2. Table IV shows that NSTSC's accuracy is $0.42\%$ lower than HCTE2, but NSTSC ranks first more often than the other systems, which can be attributed to the logical properties of wSTL. Essentially, wSTL formulas classify data into two classes, meaning NSTSC is conceptually a binary classifier. Hence NSTSC achieves a better performance on the binary tasks. The multiclass classification is accomplished by building a DT using binary-type classifiers. Such method of constructing a multiclass classifier may impact the performance of the model on multiclass classification tasks. While the above results show NSTSC's potential for TSC, its major strength is providing easily understandable and readable logical formulas to users, which is especially meaningful in practical applications. However, the other models do not possess such attributes.

TABLE II
PERFORMANCE OF ALL TSC MODELS ON THE WOUND HEALING DATA.

| Model | Accuracy | | Recall | | Precision | | F1 | |
|---|---|---|---|---|---|---|---|---|
| | mean | std | mean | std | mean | std | mean | std |
| FCN | 76.99 | 5.03 | 77.66 | 6.63 | 80.68 | 10.16 | 0.7558 | 0.0741 |
| ResNet | 57.47 | 10.54 | 61.00 | 12.18 | 55.30 | 16.61 | 0.5449 | 0.1285 |
| MLP | 23.29 | 0.92 | 20.00 | 0 | 4.66 | 0.18 | 0.0755 | 0.0024 |
| **NSTSC** | **83.85** | 9.42 | **84.17** | 8.99 | **86.33** | 7.89 | **0.8327** | 0.1001 |

TABLE III
AVERAGE CLASSIFICATION ACCURACY OF NSTSC AND SOTA MODELS ON THE UCR TIME-SERIES DATA ARCHIVE.

| Model | FCN | ResNet | ST | BOSS | HCTE | HCTE2 | ROCKET | CHIEF | **NSTSC** |
|---|---|---|---|---|---|---|---|---|---|
| Avg accuracy | 80.91 | 82.47 | 82.23 | 81.12 | **84.71** | 86.12 | 85.07 | 84.78 | 84.66 |
| No. rank 1st | 12 | 10 | 9 | 11 | 12 | **25** | 15 | 20 | **24** |

TABLE IV
AVERAGE CLASSIFICATION ACCURACY OF NSTSC AND SOTA MODELS ON THE BINARY DATASETS IN THE UCR ARCHIVE.

| Model | FCN | ResNet | ST | BOSS | HCTE | HCTE2 | ROCKET | CHIEF | **NSTSC** |
|---|---|---|---|---|---|---|---|---|---|
| Avg accuracy | 86.11 | 87.57 | 86.43 | 85.98 | 88.09 | **89.49** | 88.83 | 88.67 | **89.07** |
| No. rank 1st | 6 | 4 | 4 | 7 | 5 | 6 | 7 | 8 | **11** |

TABLE V
CLASSIFICATION ACCURACY OF THE ABLATION STUDY ON THE UCR TIME-SERIES DATA ARCHIVE.

| Ablation Study Setting | No raw data | No spectral data | No derivative data | Only raw data |
|---|---|---|---|---|
| Avg accuracy | 79.13 | 79.61 | 79.85 | 79.36 |

*2) Interpretability Analysis:* The "BeetleFly" dataset in the UCR TSD archive is selected to demonstrate the interpretabil-

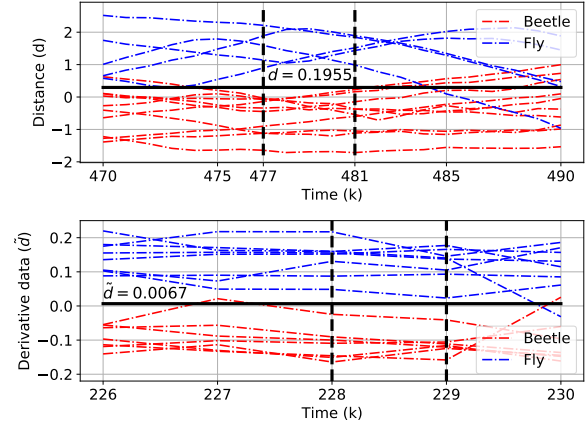

Fig. 6. Raw and derivative data of the "BeetleFly" dataset and decision rules in $\phi^{r2}$ (top) and $\phi^{d2}$ (bottom).

ity, which includes Beetle and Fly classes. The formula for the Fly data is $\phi = (\phi_{f1})^{0.0029} \land (\phi_{f2})^{\mathbf{0.9971}}$, where $\phi_{f1} = (\lozenge_{[0,531]}^{\boldsymbol{w}^{r1}} \square_{[0,531]}^{\boldsymbol{w}^{r2}} \pi^{r1}) \land (\lozenge_{[0,531]}^{\boldsymbol{w}^{s1}} \square_{[0,531]}^{\boldsymbol{w}^{s2}} \pi^{s1}) \land (\lozenge_{[0,531]}^{\boldsymbol{w}^{d1}} \square_{[0,531]}^{\boldsymbol{w}^{d2}} \pi^{d1})$, $\phi_{f2} = (\square_{[0,531]}^{\boldsymbol{w}^{r}} \pi^{r2})^{\mathbf{0.6931}} \land (\square_{[0,531]}^{\boldsymbol{w}^{s}} \pi^{s2})^{0.018} \land (\square_{[0,531]}^{\boldsymbol{w}^{d}} \pi^{d2})^{\mathbf{0.2890}}$, and $\pi^{ri}, \pi^{si}, \pi^{di}, i = 1, 2$ are atomic predicates describing raw data ($d$), spectral data ($\hat{d}$), and derivative data ($\tilde{d}$), respectively. Notice that the subformula $\phi_{f1}$ has an associated weight of 0.0029, which is much smaller than the 0.9971 of $\phi_{f2}$, indicating that $\phi_{f1}$ is insignificant in classifying the data. To validate this, we only use $\phi_{f2}$ to make prediction of the test data. The accuracy remains the same after dropping $\phi_{f1}$, which also demonstrates the property of impact ordering. Hence $\phi_{f2}$ is utilized for interpretability analysis. The weights of $\phi_{f2}$ indicate that the raw data and the derivative data are more important. As a result, $\pi^{r2} = (d \geq 0.1955), \pi^{d2} = (\tilde{d} \geq 0.0067)$. If we consider the major subformulas, $\phi$ reads as "The distance is larger than 0.1955 during 477 to 481 and the derivative is larger than 0.0067 during 228 to 229". The top-5 weights for $\boldsymbol{w}^r$ are $[w^r_{477}, w^r_{478}, w^r_{479}, w^r_{480}, w^r_{481}] = [0.005, 0.08, 0.824, 0.041, 0.01]$. The decision rule from $\pi^{r2}$ on the raw data is shown in Fig. 6 (top). We could observe during 477 to 481, the Fly data's distance is above 0.1955, and the Beetle data's distance is below 0.1955. The top-2 weights for $\boldsymbol{w}^d$ are $[w^d_{228}, w^d_{229}] = [0.998, 0.001]$. The decision rule from $\pi^{d2}$ on the derivative data is shown in Fig. 6 (bottom). We could observe during 228 to 229, the derivative of the Beetle data is below 0.0067, and the derivative of the Fly data is above 0.0067. This demonstrates the patterns from NSTSC can interpret the characteristics of data.

### C. Ablation Study

This subsection investigates the necessity of multi-view data representation in the TSC process. The node phase classifier and tree phase classifier are obviously necessary. To demonstrate the necessity of each module, we exclude it and show the degradation of the model performance. First, we study the effect of each data representation by keeping the other two representations. Second, we only use the raw data to evaluate NSTSC. Table V shows the results of the ablation study on the UCR dataset. Specifically, we have the following findings.

(1) Removing the raw, the spectral, or the derivative data degrades the average accuracy to 79.13%, 79.61%, 79.85%, respectively. This validates the multi-view data representation facilitates the learning process.

(2) The average accuracy of NSTSC with raw data only is 79.36%, which indicates using single raw data representation is still effective for TSC.

### D. Interpretability Comparison with SAX-VSM

Existing interpretable TSC models, such as Symbolic Aggregate Approximation and Vector Space Model (SAX-VSM) [35], exploits subsequences of TSD as features and finds the discriminatory subsequences to represent a class. Nevertheless, it cannot provide a human-readable formula that is intuitively interpretable. By contrast, NSTSC can both identify the discriminatory subsequences and provide an interpretable formula analogous to natural language. The "GunPoint" dataset is utilized to validate this statement. For the "Gun" class, an actor first moves his hand above a hip-mounted holster and then moves his hand down to grasp the gun and moves his hand up to the shoulder level. For the "Point" class, the actor directly moves his finger up to the shoulder level. The centroid of the actor's hand in the x-axis is tracked. Fig. 7 shows the intervals extracted by NSTSC and SAX-VSM. It is evident that intervals from NSTSC align with the subsequences identified by SAX-VSM because of the intersection. The patterns correspond to the difference between moving down to grasp the gun and directly moving up. Furthermore, if we truncate the small weights, NSTSC can provide a wSTL formula $\phi = (x(34) \leq -0.06) \vee (x(35) \leq 0.04) \vee (x(36) \leq 0.28) \vee (x(109) \leq -0.48) \vee (x(110) \leq -0.65)$, which reads in plain English as "x pos at 34 s is below -0.06 or x pos at 35 s is below 0.04 or x pos at 36 s is below 0.28 or x pos at 109 s is below -0.48 or x pos at 110 s is below -0.65", while SAX-VSM cannot provide such human-readable formulas. Note Table III excludes the comparison with SAX-VSM as the accuracies are not available in the original paper.

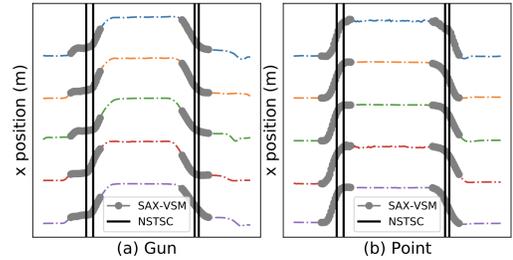

Fig. 7. Characteristic patterns in NSTSC and SAX-VSM.

## VII. CONCLUSION

In this study, we propose a neuro-symbolic model called NSTSC to classify TSD. We propose a novel weighted semantics for STL to quantify the truth degree of a wSTL formula over TSD. The NSTSC is designed by integrating the wSTL and the neural network, where each neuron has an associated symbolic expression in a wSTL formula such that the NSTSC is differentiable and can be expressed as a human-readable statement. A decision tree-based classifier is designed to learn formula structures and classify multiclass TSD. We apply the neuro-symbolic model on two groups of time-series datasets to demonstrate that it can achieve comparable performance with state-of-the-art models. Furthermore, the interpretability analysis demonstrates that NSTSC can produce logical formulas that are easy to understand and match with domain knowledge, which is a property that existing models do not possess.

## APPENDIX

### A. Conversion from a Tree Classifier to a Formula

Traversing the tree from the root node to a leaf node gives a path and a corresponding subformula describing the class of data in the leaf node. The wSTL formula conversion procedures are presented in Algorithm 3.

---

**Algorithm 3** Tree Classifier to Formulas - *TCTF()*

---

**Input:** A tree phase classifier $\mathcal{T}$ learned via Algorithm 2
**Output:** wSTL formulas for $C$ classes, $\Phi = \{\phi_1, ..., \phi_C\}$
1: Initialize a subformula list for each data class, $\boldsymbol{\phi_c} = []$
2: **for** each path $\mathcal{P}$ in $\mathcal{T}$ **do**
3:   **for** each Node $n$ in $\mathcal{P}$ **do**
4:     **if** Node $n$ is the root node **then**
5:       $\phi_{pt} = \emptyset$
6:     **else if** Node $n$ is a left child node **then**
7:       Set $\phi_{pt} = \phi_{pt} \wedge \phi^{n-1,*}$
8:     **else if** Node $n$ is a right child node **then**
9:       Set $\phi_{pt} = \phi_{pt} \wedge \neg\phi^{n-2,*}$
10:     **end if**
11:   **end for**
12:   Append $\phi_{pt}$ into $\boldsymbol{\phi_c}$ for class $c$ represented by the leaf node of $\mathcal{P}$
13: **end for**
14: **for** $c = 1, ..., C$ **do**
15:   Set $\phi_c = \emptyset$
16:   **for** each subformula $\phi_c^s$ in $\boldsymbol{\phi_c}$ **do**
17:     $\phi_c \leftarrow \phi_c \vee \phi_c^s$
18:   **end for**
19:   $\Phi \leftarrow \Phi \cup \phi_c$
20: **end for**
21: **return** $\Phi$